\documentclass[letterpaper]{article} 
\usepackage{aaai2026}  
\usepackage{times}  
\usepackage{helvet}  
\usepackage{courier}  
\usepackage[hyphens]{url}  
\usepackage{graphicx} 
\usepackage{natbib}  
\usepackage{caption} 
\usepackage{algorithm}
\usepackage{algpseudocode}
\usepackage{bm}

\usepackage{booktabs}    
\usepackage{multirow}    
\usepackage{amsmath}      
\usepackage{amssymb}      
\usepackage{xcolor}
\usepackage{cleveref}
\usepackage{colortbl} 

\usepackage{newfloat}
\usepackage{listings}
\DeclareCaptionStyle{ruled}{labelfont=normalfont,labelsep=colon,strut=off} 
\floatstyle{ruled}
\newfloat{listing}{tb}{lst}{}
\floatname{listing}{Listing}
\title{Rethinking the Spatio-Temporal Alignment of End-to-End 3D Perception}
\author{
    Xiaoyu Li\textsuperscript{\rm 1}\equalcontrib, 
    Peidong Li\textsuperscript{\rm 2,3}\equalcontrib, 
    Xian Wu\textsuperscript{\rm 1},
    Long Shi\textsuperscript{\rm 1},
    Dedong Liu\textsuperscript{\rm 1},\\
    Yitao Wu\textsuperscript{\rm 1}, 
    Jiajia Fu\textsuperscript{\rm 1}, 
    Dixiao Cui\textsuperscript{\rm 4}, 
    Lijun Zhao\textsuperscript{\rm 1}\thanks{Corresponding author.}, 
    Lining Sun\textsuperscript{\rm 1}
} 
\affiliations{
    \textsuperscript{\rm 1}State Key Laboratory of Robotics and System, Harbin Institute of Technology, Harbin, China\\
    \textsuperscript{\rm 2}IROOTECH, Guangzhou, China \\
    \textsuperscript{\rm 3}Wolf 1069b, Guangzhou, China\\
    \textsuperscript{\rm 4}Independent Researcher\\

}

\usepackage{bibentry}

\begin{document}

\maketitle

\begin{abstract}
Spatio-temporal alignment is crucial for temporal modeling of end-to-end (E2E) perception in autonomous driving (AD), providing valuable structural and textural prior information.
Existing methods typically rely on the attention mechanism to align objects across frames, simplifying the motion model with a unified explicit physical model (constant velocity, etc.). 
These approaches prefer semantic features for \textcolor{black}{implicit} alignment, challenging the importance of \textcolor{black}{explicit} motion modeling in the traditional perception paradigm. 
However, variations in motion states and object features across categories and frames render this alignment suboptimal.
To address this, we propose HAT, a spatio-temporal alignment module that allows each object to adaptively decode the optimal alignment proposal from multiple hypotheses without direct supervision.
Specifically, HAT first utilizes multiple explicit motion models to generate spatial anchors and motion-aware feature proposals for historical instances. 
It then performs multi-hypothesis decoding by incorporating semantic and motion cues embedded in cached object queries, ultimately providing the optimal alignment proposal for the target frame.
\textcolor{black}{On nuScenes, HAT consistently improves 3D temporal detectors and trackers across diverse baselines.
It achieves state-of-the-art tracking results with 46.0\% AMOTA on the test set when paired with the DETR3D detector.
In an object-centric E2E AD method, HAT enhances perception accuracy (+1.3\% mAP, +3.1\% AMOTA) and reduces the collision rate by 32\%.
When semantics are corrupted (nuScenes-C), the enhancement of motion modeling by HAT enables more robust perception and planning in the E2E AD.}
\end{abstract}

\begin{links}
    \link{Code}{https://github.com/lixiaoyu2000/HAT}
\end{links}

\section{Introduction}
\label{sec:intro}

\begin{figure}[t]
      \centering
      \includegraphics[width=\linewidth]{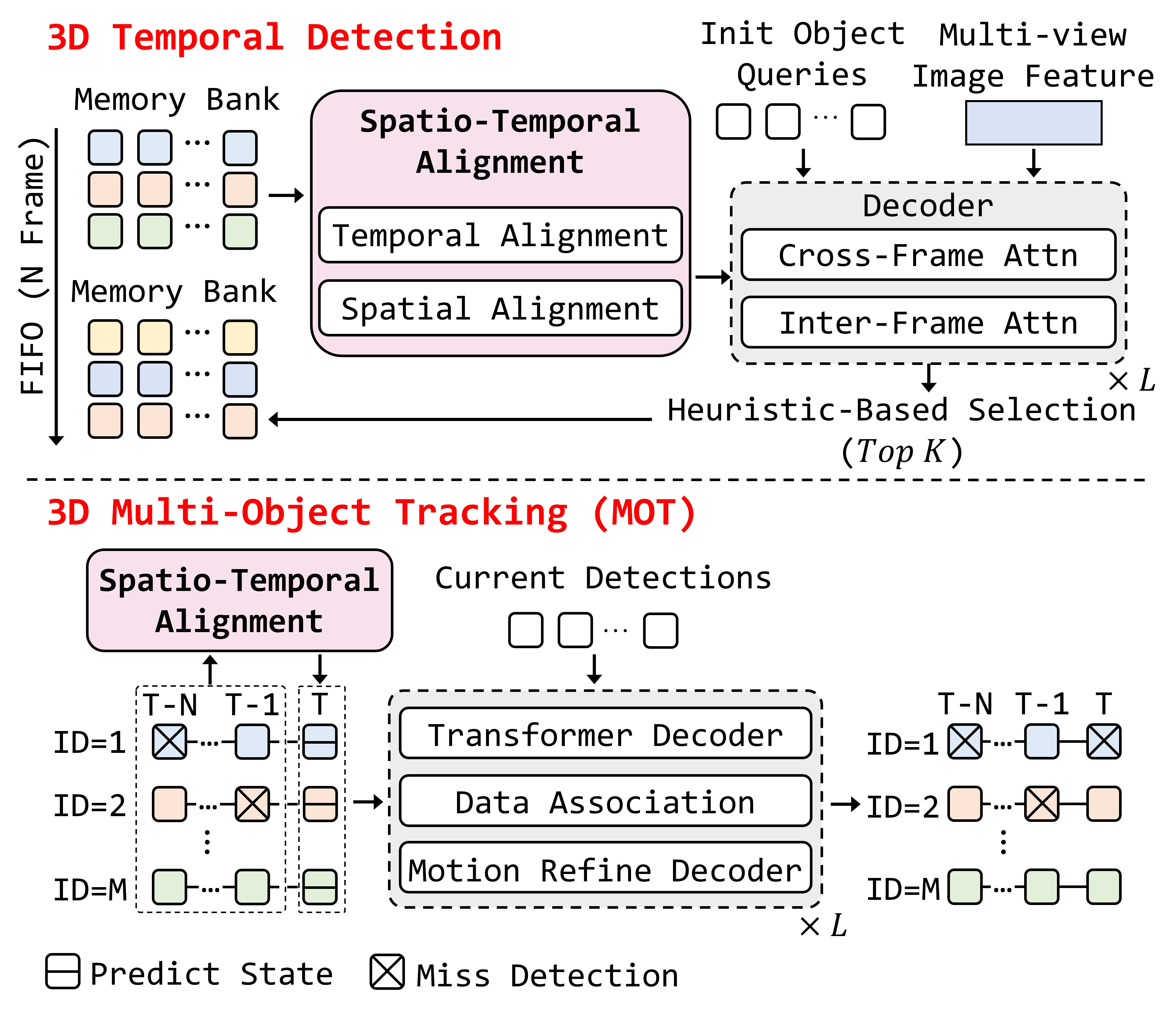}
      \caption[l]{
      The STA module in detection and tracking.
      }
    \label{fig:inro}
\end{figure}

In the autonomous driving, multi-camera 3D temporal detection~\cite{lin2023sparse4d, liu2023sparsebev, wang2023exploring, tang2025simpb, DETR3D} and multi-object tracking~\cite{pftrack, mutr3d, ding2024ada, dq-track, lin2023sparse4d, doll2023star} (MOT) tasks converge through query-based data-stream.
They form an integrated front-end for vision-based E2E perception systems.
As illustrated in \cref{fig:inro}, these methods employ a memory mechanism to store instance-level information from adjacent frames. 
This cached information serves as helpful priors of the target frame, enhancing the performance of 3D perception from temporal modeling.
To further bridge the representation gap across time and space, the Spatio-Temporal Alignment (STA) module is introduced, aligning the features and anchors of historical instances to the target frame.


The STA module is equally critical in modularized temporal-related methods~\cite{ding20233dmotformer, li2023camo}, typically implemented via auto-regressive filters (the Kalman Filter~\cite{kalman1960new}, etc.).
Approaches like~\cite{li2023poly, genovese2001interacting} improve tracking by enforcing more plausible state transitions through refined motion models.
However, identifying the optimal alignment proposal for each object in these methods requires manual hyperparameter tuning and is prone to overfitting specific motion patterns.
Paradoxically, recent E2E methods tend to oversimplify motion modeling, ignoring these prior advancements.
These methods prefer to use query propagation to perform feature alignment and adopt a single hypothesis (constant velocity, etc.) for motion compensation, favoring inter-frame alignment in latent space~\cite{wang2023exploring} rather than in explicit motion space.
This conflicts with fundamental facts that object motion is \textit{category-specific} and \textit{time-variant}, as revealed by~\cite{li2023poly} and \cref{fig:pts_weight}.
Consequently, single-hypothesis motion modeling in current E2E perception fails to capture the diverse inter-frame transformation patterns exhibited by past instances.
In contrast, we argue that the propagated query contains valuable but underutilized motion cues.
These cues can be leveraged to distinguish and construct the most appropriate structural and textural prior information for corresponding objects.
These observations motivate a central question:
\textbf{\textit{How can E2E perception integrate the strengths of advanced STA modules of stacked perception without inheriting their brittleness?}}


To this end, we propose HAT, a multiple \textbf{H}ypotheses sp\textbf{A}tio-\textbf{T}emporal alignment module that adaptively decodes the optimal alignment proposal for each object.
Specifically, we first utilize multiple predefined explicit motion models to independently generate spatially compensated anchors and motion-aware feature proposals for historical instances. 
This multi-hypothesis generator increases the diversity of alignment proposals, effectively modeling multiple possible states of an object while ensuring performance limits.
Inspired by~\cite{liu2023sparsebev, adamixing}, we then introduce an adaptive decoder that decodes these hypotheses with dynamic weights derived from the queries.
The decoder treats historical instances as guides and generates proposals as sample points in the 3D motion space.
Finally, the fused anchors and motion-aware features are mixed through a motion refinement layer and a Feed-Forward Network (FFN) to enhance their representations.
The refined anchors and features are treated as optimal spatio-temporal alignment proposals and sent to the corresponding task head.

On the large-scale dataset nuScenes~\cite{caesar2020nuscenes}, we integrate HAT into various multi-view 3D temporal detectors and 3D trackers.
Without direct supervision, HAT demonstrates strong cross-task generalization and effectiveness.
On the validation set, it yields average gains of +0.7\% NDS and +0.6\% mAP for detection, and +1.3\% MOTA and +1.0\% AMOTA for tracking.
Combining with ADA-Track~\cite{ding2024ada}, \textbf{we achieve state-of-the-art on the test set with 46.0\% AMOTA} among trackers paired with DETR3D~\cite{DETR3D}.
Furthermore, HAT positively impacts system-level performance by improving 3D perception and reducing the collision rate in several object-centric E2E AD methods.
Its robustness is further verified on the challenging nuScenes-C benchmark~\cite{dong2023nusc_c}, which introduces deliberate semantic corruptions. 
Our extensive experiments demonstrate that motion modeling remains a crucial component in E2E 3D perception, alongside semantic cues.
The primary contributions of this work are:

\begin{itemize}
\item We propose HAT, a plug-and-play spatio-temporal alignment module that can be seamlessly integrated into various object-centric methods in the E2E AD system.
\item HAT introduces a novel explicit-implicit mixing alignment module that enhances the robustness of the E2E AD system, while overcoming the manual intervention inherent in the stacked AD system.
\item HAT achieves promising improvements when integrated into existing query-based 3D temporal detectors, trackers and even E2E AD methods, serving prior information.


\end{itemize}

\section{Related Work}

\textbf{Multi-Camera 3D Temporal Detection} initially uses Bird's-Eye-View (BEV) features~\cite{bevformer} for temporal modeling.
Recent methods instead adopt object-centric temporal propagation to reduce computational overhead.
These methods typically comprise memory bank, STA module, and transformer decoder.
The memory bank retains the $K$ most confident instances from recurrent~\cite{tang2025simpb} or adjacent frames~\cite{wang2023exploring}. 
The STA module warps historical queries and anchors to the current frame, mitigating inter-frame discrepancies. 
The transformer decoder then performs cross-/inter-frame attention to retrieve current instances from the aligned past and image features~\cite{zhu2020deformable}.
Under identity-agnostic supervision, temporal detectors surprisingly capture object-level temporal consistency, driven by the query-centric attention that implicitly links similar propagated queries across frames.

\begin{figure*}[h]
      \centering
      \includegraphics[width=\linewidth]{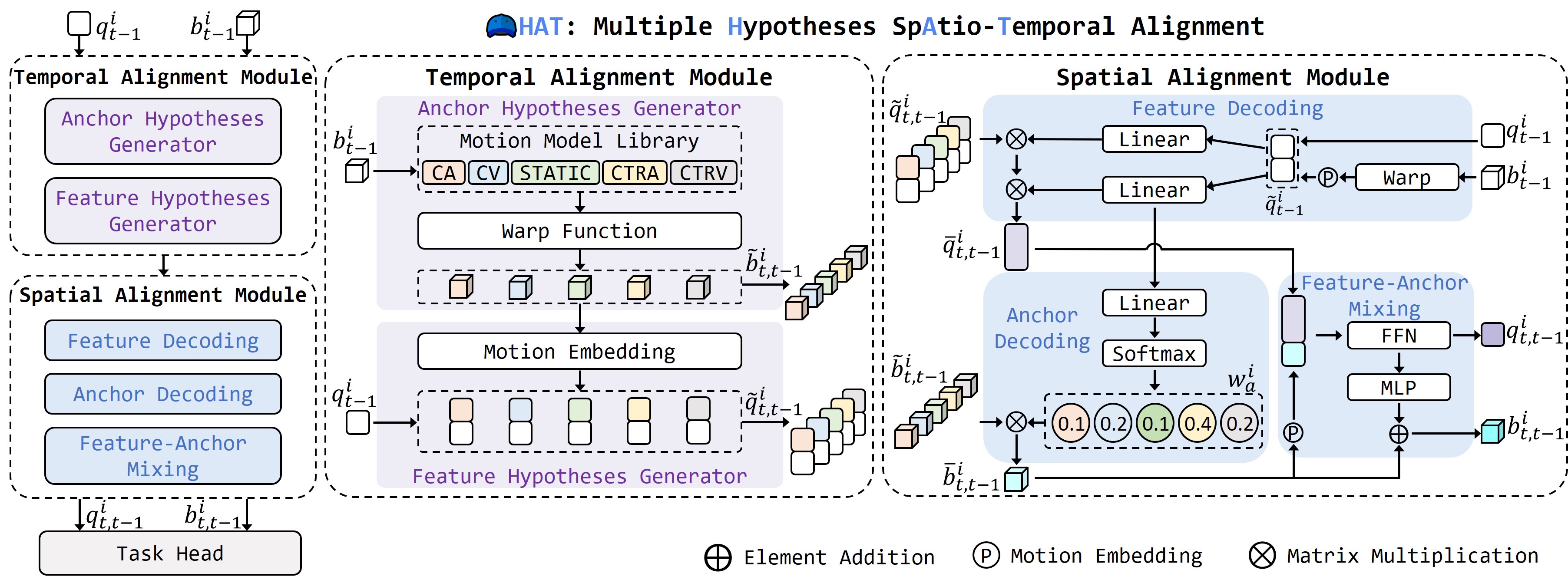}
      \caption[l]{Architecture
      of HAT. HAT takes the posterior from the previous frame as input, and outputs the prior structured (anchors) and semantic (features) information for the target frame.
      We present the alignment process for the $i$-th instance.}
    \label{fig:pipeline}
\end{figure*}

\paragraph{Multi-Camera 3D MOT} provides an explicit E2E pipeline for training detectors and trackers through collaborative query propagation. 
By modeling the consistency of instances, these trackers act as temporal-enhanced wrappers for off-the-shelf detectors~\cite{liu2022petr, DETR3D}. 
These methods typically follow the Tracking-By-Attention framework, where tracklet queries are updated from detection and image queries via attention.
Pioneering methods~\cite{mutr3d, pftrack} propagate tracklet queries while simultaneously initializing detection queries within the detection decoder.
During propagation, the STA module provides motion priors and semantic features for each tracklet, which are crucial for building and solving attention affinities.
Recent trackers~\cite{dq-track, ding2024ada} improve recall by learning an attention map between predicted tracklets and current observations.
With identity constraints, these methods achieve enhanced tracking robustness and exhibit strong detector-agnostic generalization.

\paragraph{Spatial-Temporal Alignment} functions as a prior estimator, establishing a link between past and current reasoning.
Modular tracking methods~\cite{li2023poly, li2024fast, genovese2001interacting} reduce state–observation drift in auto-regressive filters by applying STA within motion space, typically implemented through refined motion models.
However, selecting suitable models based on handcrafted rules~\cite{li2023poly} or covariance~\cite{genovese2001interacting} restricts their adaptability.
In contrast, E2E perception simplifies alignment by transferring structured priors through a unified physical model~\cite{lin2023sparse4d, ding2024ada}.
These methods prefer to leverage semantic cues to project cached features across frames in latent space, enabling data-driven temporal transitions.
Object motion and frame transformations can be incorporated as prompts~\cite{wang2023exploring, doll2023star} to enhance temporal reasoning.
Nevertheless, these approaches remain suboptimal, as a single model cannot capture the diverse motion patterns that vary across object categories and dynamic scenes.
HAT addresses this limitation by jointly modeling motion and semantics.
It leverages multiple explicit motion models to generate complementary proposals, while implicitly decoding motion and semantic cues embedded within cached instances to achieve robust and adaptive alignment.

\section{Our Approach}
\label{sec:main}

\subsection{Problem Definition}
\label{sec:overall}

The STA module aligns historical instances in the memory bank with the target timestamp. 
Without loss of generality, we consider frame $t-1$ as the historical frame and frame $t$ as the target frame.
Given the set of 3D structural anchors $B_{t-1} = \left \{ b^{i}_{t-1}, i=1 \dots K  \right \}$ and queries $Q_{t-1} = \left \{ q^{i}_{t-1}, i=1 \dots K \right \}$ for the historical instances at frame $t-1$,
$K$ denotes the number of instances stored.
STA propagates these instances to frame $t$, formulated as:
\begin{equation}
\label{eq:sta_overall}
    B_{t, t-1}, Q_{t, t-1} \longleftarrow  \operatorname{STA}\left(B_{t-1}, Q_{t-1}, \Delta t, E^{t}_{t-1}\right ), 
\end{equation}
where $B_{t, t-1}$ and $Q_{t, t-1}$ are the spatio-temporal aligned anchors and queries.
$\Delta t$ is unit time interval between adjacent frames.
$E^{t}_{t-1}= [R^{t}_{t-1}|T^{t}_{t-1}] \in \mathbb{R}^{3 \times 4} $ is the ego pose transformation matrix from frame $t-1$ to frame $t$.
$R^{t}_{t-1} \in \mathbb{R}^{3 \times 3}$ and $T^{t}_{t-1} \in \mathbb{R}^{3 \times 1}$ are the rotation and translation.

An anchor $b^{i}$ and its feature $q^{i} \in \mathbb{R}^{1 \times C}$ form the $i$-th instance.
$b^{i}$ is represented as a vector $\left [ P, D, \Theta, V \right ] \in \mathbb{R}^{1 \times 10}$, where $P = \left [ x, y, z\right ]$ denotes the 3D center.
$D = \left [w, l, h\right ]$ denotes 3D size.
$\Theta = \left [\cos{\theta}, \sin{\theta}\right ]$ is yaw vector, where $\theta$ is the overall yaw.
$V = \left [vx, vy\right ]$ is the velocity vector. 

\subsection{Overall Architecture}\label{sec:overall}

The alignment is conducted in two stages.
The Temporal Alignment Module first produces multiple motion-aware hypotheses informed by historical states and explicit motion models.
Subsequently, the Spatial Alignment Module decodes these hypotheses using the motion cues embedded in the query, selecting the optimally aligned anchor and associated feature for propagation to the current frame as prior.

\subsection{Temporal Alignment Module}\label{sec:temporal_align}

This module generates multiple anchor and feature alignment proposals based on several explicit motion models.
On the one hand, we employ modularized approaches~\cite{genovese2001interacting} to enhance robust performance, and on the other hand, we utilize query propagation to improve adaptability.
\paragraph{Multiple Anchor Hypotheses Generator.}
\textcolor{black}{Random frame sampling~\cite{tang2025simpb}, latent ID consistency modeling~\cite{sun2024sparsedrive}, and lack of map collectively restrict the generalizability of fixed-horizon motion forecasting~\cite{shi2022mtr} in the STA module.
Consequently, most methods rely on explicit motion models for anchor alignment, typically assuming a single motion hypothesis.}
In contrast, we define and store a set of well-established motion models in the Motion Model Library (MML).
Specifically, $M$ models are included: \textit{Constant Velocity (CV)}, \textit{STATIC}, \textit{Constant Acceleration (CA)}, \textit{Constant Turn Rate and Velocity (CTRV)}, \textit{Constant Turn Rate and Acceleration (CTRA)}.
\textcolor{black}{These representative models, widely adopted in transportation scenarios~\cite{schubert2008comparison}, capture a broad spectrum of motion patterns commonly observed in perception tasks, including stationary (\textit{STATIC}), linear motion (\textit{CV, CA}), and turning (\textit{CTRV, CTRA}).}
These models efficiently extrapolate anchor hypotheses using $\Delta t$ and the historical anchors $B_{t-1}$, uniformly formulated as:
\begin{equation}
    \begin{split}
    \hat{s}_{t, t-1} &= s_{t-1} + \int_{(t-1)\Delta t}^{t\Delta t} \dot{s}(\tau) d\tau  
               = s_{t-1} + \Delta s, \\
    \end{split}
    \label{anchor_predict}
\end{equation}
where $s_{t-1}$ and $\dot{s}$ are the state and its derivatives with respect to time in each anchor $b_{t-1} \in B_{t-1}$.
Specifically, these pairs correspond to: position and velocity, velocity and acceleration, yaw and yaw rate.
Notably, acceleration and yaw rate are unobservable in existing methods. 
Inspired by the human pose estimation task~\cite{zeng2022smoothnet}, we use a multi-layer perception (MLP) to decode these states from the instance feature $q_{t-1}$.
\textcolor{black}{Our approach to training acceleration is grounded in system observability, a principle ensuring that unobserved states (acceleration) can be inferred from measured states (position and velocity).}
$\hat{s}_{t, t-1}$ is the object-motion compensation state, expressed in the $t-1$ ego reference frame.
${\Delta s}$ represents the incremental changes.


Multiple anchor hypotheses are derived from the model-specific $\Delta s$. 
For example, for the position $x$, $\Delta s$ is 0 and $vx \times \Delta t$ in \textit{STATIC} and \textit{CV}. 
We perform the transformation of the $m$-th motion model for all historical instances $B_{t-1}$, obtaining $\hat{B}_{t, t-1}^{m}$.
A linear reference frame transformation is then applied to each instance $\hat{b}_{t, t-1}^{m} \in \hat{B}_{t, t-1}^{m}$, aligning these anchors as multiple candidate priors for frame $t$:
\begin{equation}
    ({\tilde{b}}_{t, t-1}^{m})^{T} = \textbf{R}^{t}_{t-1}({\hat{b}}_{t, t-1}^{m})^{T} + \textbf{T}^{t}_{t-1},
    \label{ego_warp}
\end{equation}
where,
\begin{equation}
\textbf{R}^{t}_{t-1}= \operatorname{Diag}(R^{t}_{t-1}, I_{3\times3},R^{t}_{t-1}[:2, :2], R^{t}_{t-1}[:2, :2]),
\end{equation}
\begin{equation}
\textbf{T}^{t}_{t-1} =
\begin{bmatrix}
(T^{t}_{t-1})^{T}, O_{1\times7}
\end{bmatrix}^T,
\end{equation}
where $\operatorname{Diag}(\cdot)$ is the diagonalize matrix operation.
$\textbf{R}^{t}_{t-1} \in \mathbb{R}^{10 \times 10}$ and $\textbf{T}^{t}_{t-1} \in \mathbb{R}^{10 \times 1}$ are augmented extrinsic matrices.
$O$ and $I$ are the zero and identity matrix.
Correspondingly, $\tilde{B}_{t, t-1}^{m} \in \mathbb{R}^{K\times10}$ is defined as the set of ${\tilde{b}}_{t, t-1}^{m}$.
We stack $\tilde{B}_{t, t-1}^{m}$ of every model, and generate the multiple anchor hypotheses as $\tilde{B}_{t, t-1} \in \mathbb{R}^{K \times M \times 10}$.

\paragraph{Multiple Feature Hypotheses Generator.}
Traditional approaches manually select the optimal anchor proposal from $\tilde{B}_{t, t-1}$. 
In contrast, we generate motion embeddings for each proposal, allowing the network to adaptively select the most appropriate anchor hypothesis.
Given that the primary variations in $\tilde{B}_{t, t-1}$ occur in position, velocity, and yaw, we utilize the state-decoupled encoder in~\cite{lin2023sparse4d} to extract motion embeddings from $\tilde{B}_{t, t-1}$, formulated as:
\begin{equation}
    \tilde{Q}_{t, t-1}' = \operatorname{Cat}(\Phi_{P}(\tilde{P}), 
    \Phi_{D}(\tilde{D}), \Phi_{\Theta}(\tilde{\Theta}), \Phi_{V}(\tilde{V})), 
    \label{motion_embed_encode}
\end{equation}
where $\operatorname{Cat}(\cdot)$ denotes concatenation. $\Phi_{P}$, $\Phi_{D}$, $\Phi_{\Theta}$, and $\Phi_{V}$ are MLPs that encode the distinct motion states: center hypotheses $\tilde{P}$, size hypotheses $\tilde{D}$, yaw hypotheses $\tilde{\Theta}$, and velocity hypotheses $\tilde{V}$, respectively. 
Motion embedding $\tilde{Q}_{t, t-1}' \in \mathbb{R}^{K \times M \times C}$ captures information from a single frame.
We then concatenate $\tilde{Q}_{t, t-1}'$ with the propagated query $Q_{t-1}$ to incorporate temporal object characteristics: 
\begin{equation}
    \tilde{Q}_{t, t-1} = \operatorname{Cat}(\tilde{Q}_{t, t-1}', 
    Q_{t-1}),
    \label{feature_encode}
\end{equation}
where $\tilde{Q}_{t, t-1} \in \mathbb{R}^{K \times M \times 2C}$ denotes multiple motion-aware feature hypotheses corresponding to $\tilde{B}_{t, t-1}$. 
$\tilde{Q}_{t, t-1}$ highlights inter-hypothesis variability while capturing the consistency between historical instances and each hypothesis.

\subsection{Spatial Alignment Module}\label{sec:spatial_align}
The STA module in the existing methods follows the Single-Input, Single-Output (SISO) rule.
In this section, we decode the single optimal hypothesis from $\tilde{B}_{t, t-1}$ and $\tilde{Q}_{t, t-1}$.
A key insight that we treat $\tilde{B}_{t, t-1}$ and $\tilde{Q}_{t, t-1}$ as the sampling points and features for historical instances in 3D motion space, analogous to the 2D sampling mechanism.
Inspired by~\cite{liu2023sparsebev, adamixing}, we introduce an adaptive multiple hypotheses decoder that leverages dynamic weights derived from the propagated instances. 
This decoder consists of three key components: feature decoding, anchor decoding, and feature-anchor mixing.

\begin{table*}[ht]
  \centering
  \small
  \begin{center}
  {
      \tabcolsep=5pt
      \begin{tabular}{c|c|c|cc|cc|cc}
        \toprule
        \multirow{2}{*}{} & \multirow{3}{*}{Stage} & \multirow{3}{*}{Method} & \multicolumn{2}{c|}{Detection} & \multicolumn{2}{c|}{Tracking} & \multicolumn{2}{c}{Planning} \\
        \cmidrule(lr){4-5} \cmidrule(lr){6-7} \cmidrule(lr){8-9} 
         & & & mAP$\uparrow$ & NDS$\uparrow$ & AMOTA$\uparrow$ & MOTA$\uparrow$ & L2 (m)$\downarrow$ & CR (\%) $\downarrow$ \\
        \midrule
        \multirow{7}{*}{\rotatebox{90}{nuScenes}} & \multirow{2}{*}{1st} & SparseDrive$\dagger$  & 41.9  & 53.0    & 38.2             & 35.5  & - & -\\
         & &  SparseDrive-HAT                & \textbf{42.1} \textit{\textbf{{\scriptsize (+0.2)}}}  & \textbf{53.1} \textit{\textbf{{\scriptsize (+0.1)}}}   & \textbf{40.0} \textit{\textbf{{\scriptsize (+1.8)}}}  & \textbf{37.2} \textit{\textbf{{\scriptsize (+1.7)}}}   & - & -\\
        \cmidrule(lr){2-9}
         & \multirow{4}{*}{2nd} & SparseDrive$\dagger$  & 41.2  & 52.2    & 36.9             & 34.2  & 0.63 & 0.123 \\
         & & SparseDrive-HAT                & \textbf{42.5} \textit{\textbf{{\scriptsize (+1.3)}}}  & \textbf{53.1} \textit{\textbf{{\scriptsize (+0.9)}}}  & \textbf{40.0} \textit{\textbf{{\scriptsize (+3.1)}}}  & \textbf{36.7} \textit{\textbf{{\scriptsize (+2.5)}}}  & \textbf{0.60} \textit{\textbf{{\scriptsize (-0.03)}}}  & \textbf{0.084} \textit{\textbf{{\scriptsize (-32\%)}}} \\
         \cmidrule(lr){3-9}
         &  & DiffusionDrive  & 41.2  & 52.2    & 37.5             & 34.8  & \textbf{0.57} & 0.080 \\
         & & DiffusionDrive-HAT                & \textbf{42.7} \textit{\textbf{{\scriptsize (+1.5)}}}  & \textbf{54.0} \textit{\textbf{{\scriptsize (+1.8)}}}  & \textbf{40.2} \textit{\textbf{{\scriptsize (+2.7)}}}  & \textbf{36.7} \textit{\textbf{{\scriptsize (+1.9)}}}   & 0.58  & \textbf{0.042} \textit{\textbf{{\scriptsize (-48\%)}}} \\ \cmidrule(lr){2-9}
         & -  & SSR~\cite{li2025navigationguidedsparsescenerepresentation} & \multicolumn{4}{c|}{-} & 0.39 & 0.06 \\
        \midrule
        \midrule
        \multirow{4}{*}{\rotatebox{90}{nuScenes-C}} & \multirow{2}{*}{2nd w/ Snow} & SparseDrive$\dagger$  & 18.9  & 34.1    & 13.1             & 14.1   & 0.74 & 0.156 \\
         & & SparseDrive-HAT                & \textbf{23.1} \textit{\textbf{{\scriptsize (+4.2)}}}  & \textbf{39.1} \textit{\textbf{{\scriptsize (+5.0)}}}   & \textbf{18.0} \textit{\textbf{{\scriptsize (+4.9)}}}  & \textbf{17.4} \textit{\textbf{{\scriptsize (+3.3)}}}  & \textbf{0.737} \textit{\textbf{{\scriptsize (-0.003)}}} & \textbf{0.122} \textit{\textbf{{\scriptsize (-22\%)}}} \\
        \cmidrule(lr){2-9}
         & \multirow{2}{*}{2nd w/ Fog} & SparseDrive$\dagger$  & \textbf{37.2}  & 49.6    & 32.6     & 30.1         & \textbf{0.61} & 0.108 \\
         & & SparseDrive-HAT                 & {37.1}  & {\textbf{50.3}} \textit{\textbf{{\scriptsize (+0.7)}}}   & \textbf{34.3} \textit{\textbf{{\scriptsize (+1.7)}}}  & {\textbf{31.7}} \textit{\textbf{{\scriptsize (+1.6)}}}   & {0.63} & \textbf{0.078} \textit{\textbf{{\scriptsize (-28\% )}}}  \\
        \bottomrule
      \end{tabular}
  }
  \end{center}
  \caption{Comparison on nuScenes and nuScenes-C validation sets for E2E AD task. 
  $\dagger$: Our reproduced results by official code.}
  \label{tab:test_ad}
\end{table*}

\paragraph{Feature Decoding.} 
Following~\cite{liu2023sparsebev, adamixing}, we first generate dynamic weights from $B_{t-1}$ and $Q_{t-1}$, which subsequently guide the fusion of multiple features.
$B_{t-1}$ is directly warped to frame $t$ through \cref{ego_warp}, resulting in $\tilde{B}_{t-1}$.
Via \cref{motion_embed_encode} and \cref{feature_encode}, $\tilde{B}_{t-1}$ is encoded as motion embedding, and concatenated with $Q_{t-1}$, yielding $\tilde{Q}_{t-1} \in \mathbb{R}^{K \times 2C}$.
We apply two linear layers $\operatorname{L}(\cdot)$ on $\tilde{Q}_{t-1}$ to obtain the dynamic weights, is described as:
\begin{equation}
    W_{c} = \operatorname{L}_{c}(\tilde{Q}_{t-1}),
    W_{f} = \operatorname{L}_{f}(\tilde{Q}_{t-1}),
    \label{weight_generator}
\end{equation}
where the parameters of $\operatorname{L}_{c}(\cdot)$ and $\operatorname{L}_{f}(\cdot)$ are shared across instances. 
$W_{c} \in \mathbb{R}^{K \times 2C \times 2C}$ and $W_{f} \in \mathbb{R}^{K \times M \times 1}$ are the weights of channel fusion and feature hypotheses fusion.
An MLP-like architecture then uses $W_{c}$ and $W_{f}$ to fuse $\tilde{Q}_{t, t-1}$:
\begin{equation}
\begin{split}
    \acute{Q}_{t, t-1} &= \sigma(\operatorname{LN}(\tilde{Q}_{t, t-1} \otimes W_{c})), \\
    \bar{Q}_{t, t-1} &= \sigma(\operatorname{LN}(W_{f} \otimes \acute{Q}_{t, t-1})),
\end{split}
    \label{eq:feature_fusion}
\end{equation}
where $\sigma(\cdot)$ is the activation function, and $\operatorname{LN(\cdot)}$ represents layer normalization.
The operator $\otimes$ denotes the generalized matrix multiplication, including dimension-aligned operations.
$\bar{Q}_{t, t-1} \in \mathbb{R}^{K \times 2C}$ is the motion-aware feature decoded from various feature hypotheses.
This feature selectively integrates motion embeddings from distinct motion models, guided by the motion cue in the propagated queries.


\paragraph{Anchor Decoding.}
Unlike the 2D sampling mechanism~\cite{adamixing, liu2023sparsebev}, which focuses solely on sampling features, the STA module necessitates decoding and propagating sampling points (multiple anchor hypotheses) to the target frame.
Inspired by posterior estimation of Interacting Multiple Model (IMM) filter~\cite{genovese2001interacting}, we implement a weighted sum approach to decode $\tilde{B}_{t, t-1}$.

In contrast to manually setting switching likelihood~\cite{genovese2001interacting}, we leverage the network to regress the weights adaptively.
Specifically, we perform a linear transformation and softmax function on the feature hypotheses weight $W_{f}$ to obtain the final weight. 
This can be characterized as:
\begin{equation}
    W_{a} = \operatorname{Softmax}(\operatorname{L}_{a}(W_{f})), 
    \label{anchor_weight}
\end{equation}
where $\operatorname{L}_{a}$ is the linear layer.
The parameters are shared across instances.
$W_{a} \in \mathbb{R}^{K \times M \times 1}$ is the anchor hypotheses weight.
We then use $W_{a}$ to perform a weighted sum:
\begin{equation}
    \bar{B}_{t, t-1} = W_{a} \otimes \tilde{B}_{t, t-1}, 
    \label{anchor_decode}
\end{equation}
where $\bar{B}_{t, t-1} \in \mathbb{R}^{K \times 10}$ is the optimal anchor proposal for adaptive decoding from $\tilde{B}_{t, t-1}$.
Although the anchor motion models in MML are explicit and fixed, our implicit decoder improves the ability to fit diverse motions.

\paragraph{Feature-Anchor Mixing.}
We perform enhancements on $\bar{Q}_{t, t-1}$ and $\bar{B}_{t, t-1}$, ultimately producing $Q_{t, t-1}$ and $B_{t, t-1}$ for frame $t$.
For the feature aspect, via \cref{motion_embed_encode}, we encode $\bar{B}_{t, t-1}$ into motion embedding.
This embedding is then concatenated with $\bar{Q}_{t, t-1}$ to obtain the enhanced motion-aware feature.
An FFN subsequently compresses the dimensions of the enhanced features to obtain the final output $Q_{t, t-1}$.
For the anchor aspect, we apply an MLP on $Q_{t, t-1}$, refining the decoded anchors $\bar{B}_{t, t-1}$ to improve quality:
\begin{equation}
B_{t, t-1} = \bar{B}_{t, t-1} + \Phi_{r}(Q_{t, t-1}),
\label{refine_anchor}
\end{equation}
where $\Phi_{r}$ is MLP for motion refinement.
$B_{t, t-1}$ and $Q_{t, t-1}$ represent the optimal anchor and feature alignment for frame $t$, which are subsequently forwarded to the task head.

\subsection{Stability Analysis}
HAT maintains baseline accuracy without direct supervision. 
The adaptive increment of aligned position $\bar{X}_{t,t-1}$ over the warped historical anchor is constrained within \(\left(R^{t}_{t-1}\min(\Delta X^{m}),\; R^{t}_{t-1}\max(\Delta X^{m})\right)\).
$\Delta X^{m}$ is the $m$-th model compensation. 
As these models are physically grounded, the constraint stabilizes $\bar{X}_{t,t-1}$.

\begin{table*}[h]
  \centering
  \small
  \renewcommand{\arraystretch}{1}
  \begin{center}
  {
    \begin{tabular}{c|c|c|c|c|cccccc}
      \toprule
      Set & Tracker        & E2E        & Detector & Backbone                         & MOTA$\uparrow$  & AMOTA$\uparrow$  & FP$\downarrow$ & FN$\downarrow$ & IDS$\downarrow$ & AMOTP$\downarrow$ \\
      \midrule
      \multirow{4}{*}{Val} 
      & ADA-Track        &  \multirow{2}{*}{\checkmark}       & \multirow{2}{*}{DETR3D} & \multirow{2}{*}{R101}                       & 34.7    & 38.4                   & 14358             & \textbf{38035}             & 839       & 1.378       \\
      & ADA-Track-HAT      &           &  &                    & \textbf{36.4} \textit{\textbf{{\scriptsize (+1.7)}}}       &  \textbf{39.7} \textit{\textbf{{\scriptsize (+1.3)}}}                   & \textbf{13100}             & 38121             & \textbf{752}        & \textbf{1.344}      \\
      \cmidrule{2-11}
      & StreamPETR       &  \multirow{2}{*}{$\times$}         & \multirow{2}{*}{StreamPETR} & \multirow{2}{*}{V2-99}                 & 46.1           & 52.6                   & \textbf{12594}             & 33380             & \textbf{742}  & 1.129            \\
      & StreamPETR-HAT      &           &  &                       & \textbf{47.0 \textit{{\scriptsize (+0.9)}}}   &  \textbf{53.3 \textit{{\scriptsize (+0.7)}} }                 & 14073             & \textbf{30946}             & 775          & \textbf{1.107}    \\
      \midrule
      \midrule
      \multirow{6}{*}{Test}
      & MUTR3D             & \multirow{3}{*}{\checkmark}          & \multirow{3}{*}{DETR3D}           & \multirow{3}{*}{V2-99}      & 24.5    & 27.0                & 15372 & 56874 & 6018  & 1.494       \\
      & PF-Track                 &          &             &      & 37.8  & 43.4                & 19048 & 42758 & \textbf{249} & 1.252         \\
      & STAR-Track                 &        &            &   & 40.6    & 43.9                & -- & -- & 607 & 1.256         \\
      \cmidrule{2-11}
      & ADA-Track                 & \multirow{2}{*}{\checkmark}          & \multirow{2}{*}{DETR3D} & \multirow{2}{*}{V2-99}        & 40.6  & 45.6                & 15699 & \textbf{39680} & 834  & 1.237        \\
      &  ADA-Track-HAT$\dagger$                                & &     &  &  \textbf{41.6}  \textit{\textbf{{\scriptsize (+1.0)}}} & \textbf{46.0} \textit{\textbf{{\scriptsize (+0.4)}}}                      & \textbf{15235} & 39799 & 850  & \textbf{1.236}\\
      \bottomrule
    \end{tabular}
  }
\end{center}
\caption{Comparison results on nuScenes validation and test sets for the tracking benchmark. E2E: end-to-end trackers that integrate detection and tracking through object queries. \textcolor{black}{$\dagger$: we re-implement the baseline as it is closed-source on the test set.}}
\label{tab:track_combined_track}
\end{table*}

\begin{table}[h]
  \centering
  \small
  \renewcommand{\arraystretch}{1}
  \begin{center}
    {
    \tabcolsep=2pt
    \begin{tabular}{c|ccccc}
      \toprule
      Detector      & NDS$\uparrow$    & mAP$\uparrow$         & mATE$\downarrow$    & mAOE$\downarrow$ & mAVE$\downarrow$ \\
      \midrule
      StreamPETR                & 57.1    & 48.2                         & 0.61             & 0.38             & 0.26            \\
      w/ HAT                  & \textbf{57.8 \textit{{\scriptsize (+0.7)}}}   &  \textbf{48.7 \textit{{\scriptsize (+0.5)}}}                     & \textbf{0.59}             & \textbf{0.37}             & \textbf{0.24}          \\
      \midrule
      Sparse4D                & 56.4  & 46.5                    & 0.54            & 0.46            & 0.22          \\
      w/ HAT                & \textbf{57.3} \textit{\textbf{{\scriptsize (+0.9)}}}  &  \textbf{47.0} \textit{\textbf{{\scriptsize (+0.5)}} }                      & \textbf{0.53}             & \textbf{0.42}             & \textbf{0.21}           \\
      
      \midrule
      SimPB            & 58.6    & 47.9                       & \textbf{0.54}             & \textbf{0.32}             & 0.22            \\
      w/ HAT                 & \textbf{59.0} \textit{\textbf{{\scriptsize (+0.4)}}}   &  \textbf{48.8} \textit{\textbf{{\scriptsize (+0.9)}}}                     & 0.55             & 0.33             & \textbf{0.21}           \\
      \bottomrule
      
    \end{tabular}
    }
\end{center}
\caption{Comparison results on nuScenes validation set for the detection benchmark. All methods use an input image resolution of 256 × 704. StreamPETR employs V2-99~\cite{vovnet} as backbone, others use R50~\cite{resnet}. }
\label{tab:detection_val}
\end{table}

\begin{table}[t]
  \centering
  \small
  \renewcommand{\arraystretch}{1}

  \begin{center}
      \tabcolsep=0.7pt
      \begin{tabular}{c|ccccc}
        \toprule
        Method  & NDS$\uparrow$ & mAP$\uparrow$   & mATE$\downarrow$    & mAOE$\downarrow$ & mAVE$\downarrow$ \\
        \midrule 
        w/o MLN* & 57.0 & 48.1                     & 0.614          & 0.381            & 0.257             \\
        MLN~\cite{wang2023exploring}     & 57.1  & 48.2                     & 0.610          & 0.375            & 0.263              \\
        LMM$\dagger$~\cite{doll2023star}  & 57.5    & 48.5                    & 0.611          & \textbf{0.367}            & \textbf{0.185}              \\

        HAT (Ours) & \textbf{57.8} & \textbf{48.7}                      & \textbf{0.593}          & 0.374            & 0.244              \\
        
        \bottomrule
      \end{tabular}
  \end{center}
\caption{Distinct STA modules in StreamPETR. $\dagger$: the use of pretraining scheme in trajectory prediction. *: we disable MLN and reproduce the results.}
\label{tab:sha_module}
\end{table}

\begin{table}[h]
\centering
\small
\renewcommand{\arraystretch}{1}
{
\tabcolsep=5pt
\begin{tabular}{c|c|c|c|c|c|c}
\toprule
\multirow{2}{*}{Method} & \multicolumn{3}{c|}{\textit{Pedestrian}} & \multicolumn{3}{c}{\textit{Bicycle}} \\
\cmidrule(lr){2-4} \cmidrule(lr){5-7}
& AP $\uparrow$ & AOE $\downarrow$ & AVE $\downarrow$ 
& AP $\uparrow$ & AOE $\downarrow$ & AVE $\downarrow$ \\ 
\midrule
MLN  & 54.3 & \textbf{0.406} & 0.302 & 48.6 & \textbf{0.655} & 0.174 \\
HAT & \textbf{55.2} & 0.414 & \textbf{0.289} & \textbf{50.5} & 0.666 & \textbf{0.139} \\
\bottomrule
\end{tabular}
}
\caption{Category-wise performance in StreamPETR.}
\label{tab: Per-category Effect of HAT}
\end{table}

\begin{table}[h]
  \centering
  \small
  \renewcommand{\arraystretch}{1}
  {
  \tabcolsep=0.6pt
  \begin{tabular}{ccccc|ccccc}
    \toprule
    \multicolumn{5}{c|}{Motion Model Library} & \multirow{2}{*}{NDS$\uparrow$} & \multirow{2}{*}{mAP$\uparrow$} & \multirow{2}{*}{mATE$\downarrow$} & \multirow{2}{*}{mAOE$\downarrow$} & \multirow{2}{*}{mAVE$\downarrow$}  \\
    \scriptsize{CV} & \scriptsize STATIC & \scriptsize CA & \scriptsize CTRA & \scriptsize CTRV & & \\
    \midrule
    \checkmark &$\times$  & $\times$  & $\times$  & $\times$   & 56.5   & 45.7       & 0.55            & \textbf{0.41}          & \textbf{0.21}             \\
    $\times$  &\checkmark       & $\times$  & $\times$  & $\times$  & 56.4  & 46.2        & 0.56            & 0.43          & 0.22             \\
    \checkmark  &\checkmark       & \checkmark  & $\times$  & $\times$  & 56.6  & 46.3        & \textbf{0.53}            & 0.53          & 0.23             \\
    \checkmark    &\checkmark     & \checkmark  & \checkmark  & \checkmark  & \textbf{57.3}  & \textbf{47.0}        & \textbf{0.53}            & 0.42          & \textbf{0.21}             \\
    \midrule
    $\times$ &$\times$  & $\times$  & $\times$  & $\times$   & 55.5      & 45.7    & 0.55            & 0.48          & 0.27             \\
    \bottomrule
  \end{tabular}
  }
  \caption{Different motion model combinations in Sparse4D.}
  \label{tab:motion_model_lib}
\end{table}

\section{Experiment}
\subsection{Dataset and Implementation Details}

\textbf{Dataset and Metrics.}
We evaluate HAT on the \textbf{nuScenes} dataset, which includes 1K scenes with 1.4M 3D annotations across multiple categories. 
It provides data from 6 cameras, 1 LiDAR, along with ego pose.
HAT is further evaluated on \textbf{nuScenes-C}~\cite{dong2023nusc_c}, which introduces visual corruption under diverse weather conditions.
For \textbf{detection}, we adopt nuScenes Detection Score (NDS) and mean Average Precision (mAP) as primary metrics, and report Average Translation/Orientation/Velocity Error (ATE, AOE, AVE).  
For \textbf{tracking}, we report Average MOT Accuracy/Precision (AMOTA, AMOTP), MOT Accuracy (MOTA), along with ID Switch (IDS), False Positives/Negatives (FP, FN).  
For \textbf{E2E AD}, we report L2 error and Collision Rate (CR).

\paragraph{Implementation Details.}
For fair comparison, HAT is integrated into open-source baselines with identical configurations.  
To thoroughly assess its generalizability within the E2E pipeline, we integrate HAT into SparseDrive~\cite{sun2024sparsedrive} and DiffusionDrive~\cite{liao2025diffusiondrive}.
In independent 3D perception tasks, we apply HAT to several temporal detectors (StreamPETR~\cite{wang2023exploring}, Sparse4D~\cite{lin2023sparse4d}, SimPB~\cite{tang2025simpb}) and 3D MOT methods (ADA-Track~\cite{ding2024ada}, StreamPETR).
Image resolutions and backbone settings are listed in \cref{tab:detection_val} and \cref{tab:track_combined_track}.  
No pretraining or direct supervision is used for HAT.  
In StreamPETR, HAT fully replaces MLN.  
SmoothNet~\cite{zeng2022smoothnet} is employed to regress acceleration and yaw rate in an unsupervised manner, with outputs constrained to $\pm 0.1$.  
Motion embeddings for anchor hypotheses are extracted via the anchor encoder from~\cite{lin2023sparse4d}.  

\subsection{Comparative Evaluation}


\paragraph{E2E Autonomous Driving.}
In \cref{tab:test_ad}, we apply HAT to a query-based E2E AD method, SparseDrive. 
HAT improves the perception performance of the entire system.
For detection, HAT improves the baseline with 1.3\% mAP and 0.9\% NDS. 
For tracking, a larger 3.1\% AMOTA and a larger 2.5\% MOTA are introduced by integrating HAT. 
Meanwhile, HAT reduces the trajectory error of both ego and objects, providing a lower CR (-32\%) to ensure safety. 
Integrating HAT into DiffusionDrive~\cite{liao2025diffusiondrive} demonstrates its stability under stronger baselines.
We also surprisingly find that integrating \textbf{HAT can prevent E2E AD from the usual decrease of perception performance when joint-training with motion \& planning tasks in the second stage}.
\textcolor{black}{Furthermore, on nuScenes-C under semantic corruption (snow), the motion cues enhanced by HAT significantly improve the robustness of perception within the E2E AD framework.
HAT yields improvements of +5.0\% NDS and +4.9\% AMOTA in detection and tracking, while reducing the collision rate by 22\%.
In terms of computational overhead, HAT introduces an additional 7ms latency per frame over the 111ms baseline, indicating practical deployability.}

\paragraph{Detection.}
As shown in \cref{tab:detection_val}, HAT enhances various 3D temporal detectors on the nuScenes validation set, demonstrating generalization and effectiveness. 
It consistently improves advanced query-based detectors, increasing NDS and mAP by (0.7\%, 0.9\%, 0.4\%) and (0.5\%, 0.5\%, 0.9\%) for StreamPETR, Sparse4D, and SimPB. 
Embedding HAT also reduces heading error and velocity error, which stem from combining multiple well-established motion models.

\paragraph{Multi-Object Tracking.}
The \cref{tab:track_combined_track} demonstrates that the improvements of combining HAT in 3D MOT are also pronounced.
HAT boosts the E2E tracker ADA-Track by 1.3\% in AMOTA and 1.7\% in MOTA.
Additionally, with enhanced velocity and yaw observations, HAT improves the modularized tracker StreamPETR, yielding gains of 0.7\% in AMOTA and 0.9\% in MOTA.
On the test set, combining HAT with ADA-Track \textbf{achieves state-of-the-art performance, with 46.0\% AMOTA} among trackers paired with DETR3D detector. 
HAT improves the baseline across key metrics (+0.4\% AMOTA, +1\% MOTA), demonstrating the effectiveness. 
We argue that the tracking task, which typically relies on supervision of ID consistency, particularly benefits from the optimal alignment proposal decoding.





\subsection{Ablation Studies and Qualitative Analysis}


\begin{table}[t]
  \centering
  \small
  \renewcommand{\arraystretch}{1}
  
  {
  \tabcolsep=1.3pt
  \begin{tabular}{c|cccccc}
    \toprule
     Method & MOTA$\uparrow$ & AMOTA$\uparrow$   & FP$\downarrow$ & FN$\downarrow$ & IDS$\downarrow$ & AMOTP$\downarrow$\\
    \midrule
    Baseline        & 60.7 & 71.2     & \textbf{13010}  & 19281 & \textbf{341} & \textbf{0.459}\\
    Baseline-HAT    & \textbf{60.8} & 71.2     & 13470  & \textbf{19095} & 384 & 0.523\\
    \bottomrule
  \end{tabular}
  }
  \caption{Integrating HAT in 3DMOTFormer.}
  \label{tab:instance_modeling}
\end{table}

\begin{figure*}[t]
      \centering
      \includegraphics[width=\linewidth]{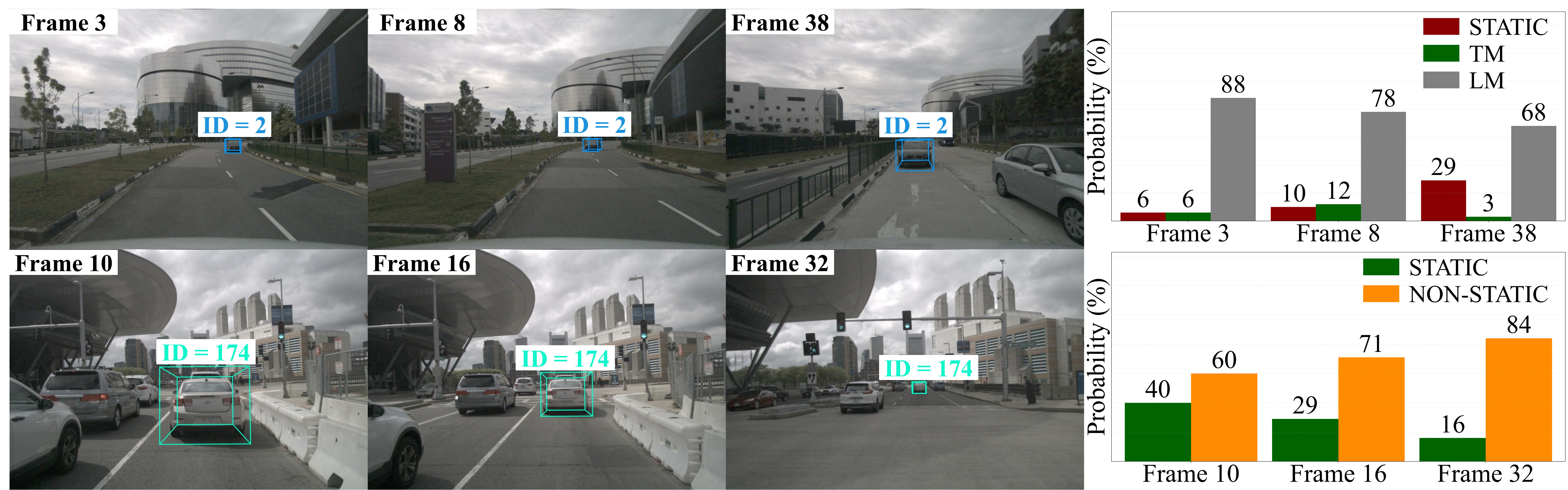}
      \caption[l]{Visualization results of the decoding weights $W_a$ for anchor hypotheses in ADA-Track-HAT on nuScenes. TM means Turning Motion model (\textit{CTRV, CTRA}), LM means Linear Motion model (\textit{CV, CA}).
      }
    \label{fig:pts_weight}
\end{figure*}

\paragraph{Effect of the STA Module.}
In~\cref{tab:sha_module}, embedding STA module generally yields a performance boost (Line 1 vs. Lines 2–4). 
This reveals the gap between the past and current, necessitating mapping in the state space.
HAT outperforms MLN, which only leverages semantic cues for implicit alignment, by 0.7\% NDS and 0.5\% mAP. 
This underscores the significance of motion cues in STA module. 
Compared to LMM, which regresses inter-frame feature projections using supervised networks, HAT employs an auto-regressive method with propagated queries. 
Furthermore, incorporating diverse proposals, HAT achieves improvements of 0.3\% NDS and 0.2\% mAP over LMM.
\textcolor{black}{In~\cref{tab: Per-category Effect of HAT}, HAT achieves notable improvements over MLN on non-rigid classes, enhancing robustness in unstructured environments.}


\paragraph{Effect of the Combination of Motion Models in MML.}
In \cref{tab:motion_model_lib}, increasing motion models in MML, based on Sparse4D, consistently enhances detection accuracy. Incorporating multiple hypotheses improves generalization and mitigates overfitting, yielding significant improvements in both NDS and mAP compared to a single hypothesis (Line 1-2). 
Compared with Line 3 without modeling the transition of yaw angle, mAOE has been greatly reduced by including \textit{CTRV} and \textit{CTRA} (Line 4), while the mAP is also improved by +0.7\%.
\textcolor{black}{We further evaluate explicit motion models by using only implicit query compensation (MLN) in MML.
Increased mAVE and mAOE highlight the importance of physical models for accurate motion estimation.}

\paragraph{Effect of the Representation of Historical Instance.}
To investigate multi-hypothesis decoding, we integrate HAT into 3DMOTFormer~\cite{ding20233dmotformer}, which propagates information via 3D decoded anchors. 
We first encode the recurrent anchors as motion embeddings. 
As shown in \cref{tab:instance_modeling}, HAT yields marginal improvement. 
We attribute this to insufficient motion cues in the decoded structure, limiting effective proposal fusion. 
In contrast, temporally propagated queries provide rich semantic and motion cues, enabling better selection, as evidenced by gains in \cref{tab:detection_val} and \cref{tab:track_combined_track}.

\paragraph{Visualization.}

\begin{figure}[t]
  \centering
  \includegraphics[width=0.9\linewidth]{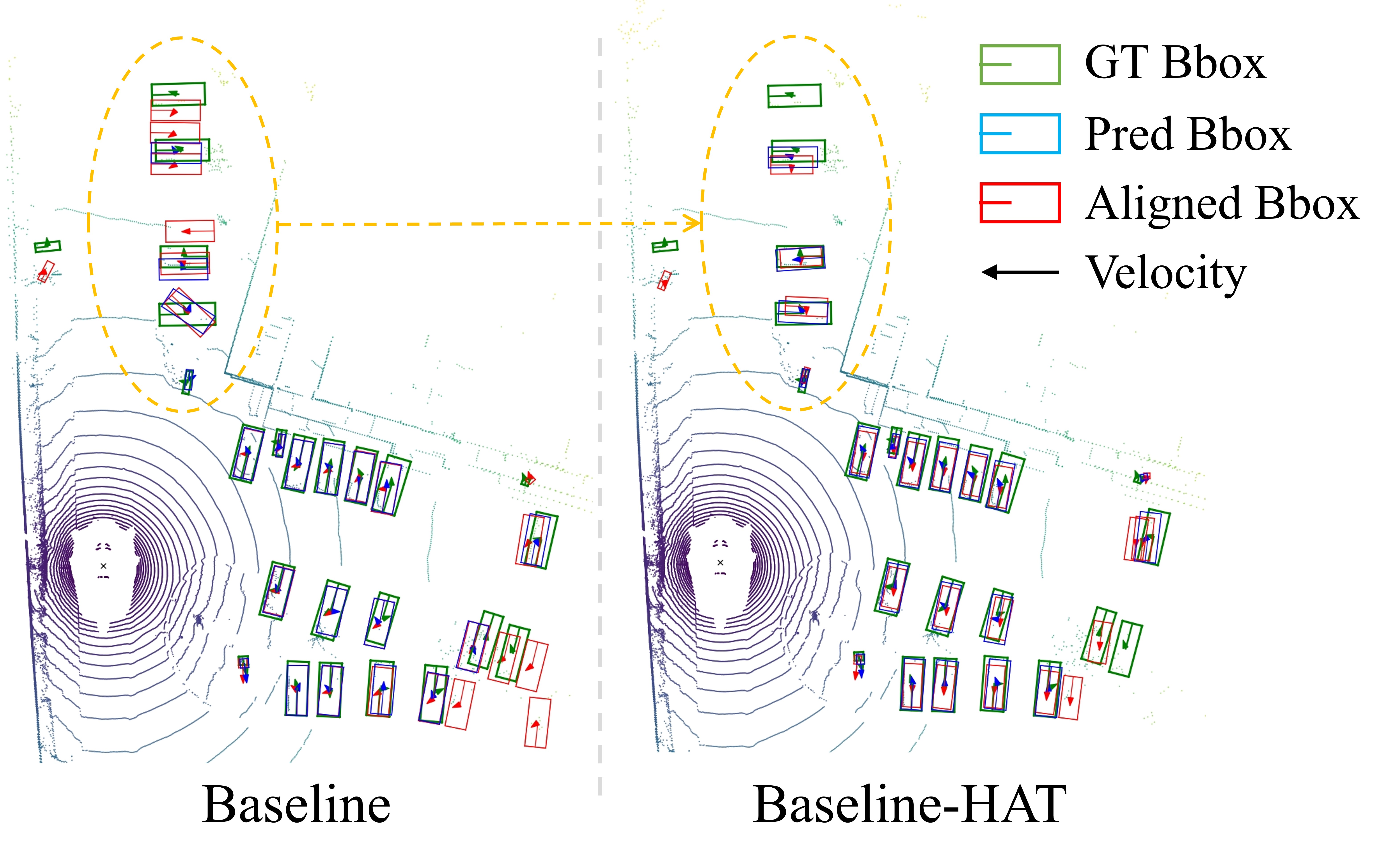}
  \caption[l]{Visualization results of Sparse4D and Sparse4D-HAT on the BEV plane on nuScenes dataset.
  The ground-truth, final prediction $B_{t}$ and temporal-aligned detection $B_{t, t-1}$ are depicted as green, blue, and red rectangles.
  }
\label{fig:detect_vis}
\end{figure}

Qualitative results demonstrate the efficacy of HAT. 
As shown in \cref{fig:detect_vis}, HAT reduces prior FPs by enhancing the temporal consistency of object instances, improving detection accuracy. 
Furthermore, \cref{fig:pts_weight} illustrates the effectiveness of HAT in tracking by assigning greater weight to the turning-motion patterns during lane changes and to the static model during braking. 
These results demonstrate that HAT dynamically modulates the contribution of explicit models based on the context, enhancing alignment.



\section{Conclusion}
In this paper, we review the contribution of motion and semantic information to the STA module. 
The proposed HAT introduces a learnable STA module, adaptively decoding the optimal proposal from hypotheses extrapolated from multiple well-established motion models.
Meanwhile, with low integration overhead, HAT exhibits impressive cross-task generalization on various AD tasks.
Our extensive experiments demonstrate that motion modeling still plays a crucial role in end-to-end 3D perception, alongside semantic cues.

\paragraph{Limitation.}
Nonetheless, the multi-hypothesis decoding mechanism
in HAT leverages the motion cues inherent in the temporally progressive query. 
As indicated in~\cref{tab:instance_modeling}, its effectiveness is reduced for methods that depend on decoded bounding boxes as the sole instance representation.

\section{Acknowledgements}
This work is supported by the Self-Planned Task of State Key Laboratory of Robotics (SKLRS202501E). In addition, the authors would like to express their sincere gratitude to Ziying Song for his valuable assistance.

\section{Appendix}
Our appendix contains an additional description of HAT, including the following sections:
\begin{itemize}
    \item \textbf{Algorithm Details.} We provide the algorithm workflow of HAT, the specific content of the Motion Model Library, and the stability proof of HAT.
    \item \textbf{Additional Experiments.} We conduct more comparative experiments and real-time quantification tests to evaluate the effectiveness and run-time performance of HAT.
    \item \textbf{Performance Verification.} We provide a screenshot of the evaluation results of ADA-Track-HAT on the nuScenes test set.
    \item \textbf{Qualitative Results.} We present additional visualizations based on the E2E AD task to further illustrate the effectiveness of HAT.
\end{itemize}

\label{sec:appendix}

\subsection{Algorithm Details}
\paragraph{Algorithm Workflow.}
\label{sec:algo_workflow}

\begin{algorithm}[!h]
\caption{\textbf{HAT}: Multiple \textbf{H}ypotheses Sp\textbf{A}tio-\textbf{T}emporal Alignment}
\newcommand{\selfin}{\textbf{Input:}}

\newcommand{\motionmodel}{We consider frame $t-1$ as the historical frame and frame $t$ as the target frame.}

\newcommand{\Statexx}{\Statex \hspace{-\algorithmicindent}}

\newcommand{\selfout}{\textbf{Output:}}
\begin{algorithmic}[1]

\Statex \motionmodel
\Statexx \selfin
\Statex \textbf{1) }Historical instances consist of corresponding features ${Q}_{t-1}$ and anchors ${B}_{t-1}$. 
\Statex \textbf{2) }Time interval between adjacent frames $\Delta {t}$.
\Statex \textbf{3) }Ego pose transformation matrix ${E}^{t}_{t-1}$.
\Statex \textbf{4) }Motion Model Library: MML =\{\textit{CV}, \textit{STATIC}, \textit{CA}, \textit{CTRV}, \textit{CTRA}\}.
\Statexx \selfout
\Statex Spatio-temporal aligned instances consist of the optimal features alignments ${Q}_{t,t-1}$ and anchors alignments ${B}_{t,t-1}$ at frame $t$.

\Statexx \textbf{(1) Temporal Alignment Module}
\Statexx \quad /** Multiple Anchor Hypotheses Generator **/
\State $\tilde{B}^{m}_{t,t-1} = \phi$
\For{\(\operatorname{Motion} \in MML\)}
\State $\hat{B}_{t,t-1} \gets \operatorname{Motion}(B_{t-1}) $
\State $\tilde{B}_{t,t-1} \gets \cref{ego_warp} (E^{t}_{t-1}, \hat{B}_{t,t-1})$ \# Reference Frame Transformation
\State $\tilde{B}^{m}_{t,t-1} \gets \tilde{B}^{m}_{t,t-1} \cup{\tilde{B}_{t,t-1}} $
\EndFor
\State $\tilde{B}_{t,t-1} \gets 
\operatorname{Stack}(\tilde{B}^{m}_{t,t-1})$
\Statexx \quad /** Multiple Feature Hypotheses Generator **/
\State $\tilde{Q}_{t, t-1}' \gets \cref{motion_embed_encode}(\tilde{B}_{t,t-1})$ \# Motion Embedding Encoding
\State $\tilde{Q}_{t, t-1} \gets \operatorname{Cat}(\tilde{Q}_{t, t-1}', Q_{t-1})$ 
\Statexx\textbf{(2) Spatial Alignment Module}
\Statexx \quad /** Feature Decoding **/
\State $ \tilde{B}_{t-1} \gets \cref{ego_warp} (E^{t}_{t-1}, {B}_{t-1})$
\State $\tilde{Q}_{t-1}' \gets \cref{motion_embed_encode}(\tilde{B}_{t-1})$
\State $\tilde{Q}_{t-1} \gets \operatorname{Cat}(\tilde{Q}_{t-1}', Q_{t-1})$
\State $W_{c}, W_{f} \gets \operatorname{Linear}_{c}(\tilde{Q}_{t-1}), \operatorname{Linear}_{f}(\tilde{Q}_{t-1})$
\State $\acute{Q}_{t, t-1} = \sigma(\operatorname{LN}(\tilde{Q}_{t, t-1} \otimes W_{c}))$
\State $\bar{Q}_{t, t-1} = \sigma(\operatorname{LN}(W_{f} \otimes \acute{Q}_{t, t-1}))$

\Statexx \quad /** Anchor Decoding **/
\State $W_{a} \gets \operatorname{Softmax}(\operatorname{Linear}_{a}(W_{f}))$
\State $\bar{B}_{t, t-1} = W_{a} \otimes \tilde{B}_{t, t-1},$
\Statexx \quad /** Feature-Anchor Mixing **/
\State $\bar{Q}_{t, t-1}' \gets \cref{motion_embed_encode}(\bar{B}_{t, t-1})$
\State $Q_{t,t-1} \gets \operatorname{FFN}(\operatorname{Cat}(\bar{Q}_{t, t-1}', \bar{Q}_{t, t-1}))$
\State $B_{t, t-1} = \bar{B}_{t, t-1} + \operatorname{MLP}(Q_{t, t-1})$

\end{algorithmic}
\end{algorithm}

The workflow of our proposed Multiple Hypotheses Spatio-Temporal Alignment is shown in Algorithm 1.

\paragraph{Motion Model Library.}
\label{sec:app_mml}
In this section, we elaborate on the state transition~\cref{anchor_predict} of each motion model in MML.
Notably, we assume the states $z, w, l, h$ are constant for all motion models.

\textbf{CV.} \textit{CV} is the abbreviation of \textit{Constant Velocity}.
The \textit{X-axis} position $x$ transition function is described by:
\begin{equation}
    \begin{split}
    \hat{x}_{t, t-1} &= x_{t-1} + \int_{(t-1)\Delta t}^{t\Delta t} {vx}_{t-1} d\tau  \\
               &= x_{t-1} + {vx}_{t-1} \Delta t.
    \end{split}
    \label{cv_x_predict}
\end{equation}

The \textit{Y-axis} position $y$ transition function is described by:
\begin{equation}
    \begin{split}
    \hat{y}_{t, t-1} &= y_{t-1} + \int_{(t-1)\Delta t}^{t\Delta t} {vy}_{t-1} d\tau  \\
               &= y_{t-1} + {vy}_{t-1} \Delta t.
    \end{split}
    \label{cv_y_predict}
\end{equation}

The \textit{X-axis} velocity $vx$ and \textit{Y-axis} velocity $vy$ remain consistent during state transitions.

The \textit{X-axis} yaw vector $\cos{\theta}$ and \textit{Y-axis} yaw vector $\sin{\theta}$ remain consistent during state transitions.

\textbf{STATIC.}
The \textit{X-axis} position $x$ and the \textit{Y-axis} position $y$ transition function is described by:
\begin{equation}
    \begin{split}
    \hat{x}_{t, t-1} &= x_{t-1},  \\
    \hat{y}_{t, t-1} &= y_{t-1}.  
    \end{split}
    \label{static_xy_predict}
\end{equation}

The \textit{X-axis} velocity $vx$ and \textit{Y-axis} velocity $vy$ are both zero, reflecting no movement:
\begin{equation} 
    \begin{split}
    \hat{vx}_{t, t-1} &= 0, \\
    \hat{vy}_{t, t-1} &= 0.  
    \end{split}
    \label{static_vxvy_predict}
\end{equation}

The \textit{X-axis} yaw vector $\cos{\theta}$ and \textit{Y-axis} yaw vector $\sin{\theta}$ remain consistent during static state transitions.

\textbf{CA.}
\textit{CA} is the abbreviation of \textit{Constant Acceleration}.
The \textit{X-axis} acceleration $ax_{t-1}$ and \textit{Y-axis} acceleration $ay_{t-1}$ are decoded from the instance query $q_{t-1}$.
$ax_{t-1}$ and $ay_{t-1}$ remain consistent during state transitions.

The \textit{X-axis} position $x$ transition function is described by:
\begin{equation}
    \begin{split}
    \hat{x}_{t, t-1} &= x_{t-1} + \int_{(t-1)\Delta t}^{t\Delta t} vx(\tau) d\tau \\
    &= x_{t-1} + {vx}_{t-1} \Delta t + \frac{1}{2} {ax}_{t-1} (\Delta t)^2.
    \end{split}
    \label{ca_x_compact}
\end{equation}

The \textit{Y-axis} position $y$ transition function is described by:
\begin{equation}
    \begin{split}
    \hat{y}_{t, t-1} &= y_{t-1} + \int_{(t-1)\Delta t}^{t\Delta t} vy(\tau) d\tau \\
    &= y_{t-1} + {vy}_{t-1} \Delta t + \frac{1}{2} {ay}_{t-1} (\Delta t)^2.
    \end{split}
    \label{ca_y_compact}
\end{equation}

The \textit{X-axis} velocity $vx$ transition function is described by:
\begin{equation}
    \begin{split}
    \hat{vx}_{t, t-1} &= {vx}_{t-1} + \int_{(t-1)\Delta t}^{t\Delta t} {ax}_{t-1} d\tau \\
               &= {vx}_{t-1} + {ax}_{t-1} \Delta t.
    \end{split}
    \label{ca_vx_predict}
\end{equation}

The \textit{Y-axis} velocity $vy$ transition function is described by:
\begin{equation}
    \begin{split}
    \hat{vy}_{t, t-1} &= {vy}_{t-1} + \int_{(t-1)\Delta t}^{t\Delta t} {ay}_{t-1} d\tau \\
               &= {vy}_{t-1} + {ay}_{t-1} \Delta t.
    \end{split}
    \label{ca_vy_predict}
\end{equation}

\textbf{CTRV.}
\textit{CTRV} is the abbreviation of \textit{Constant Turn Rate and Velocity}.
Notably, in both \textit{CTRA} and \textit{CTRV}, the velocity is represented as a single vector coupled with the heading angle, rather than in component form.
The combined velocity $v_{t-1}$ is defined as the Euclidean norm of $(vx_{t-1}, vy_{t-1})$.
The turning rate $\omega_{t-1}$ is decoded from the instance query $q_{t-1}$, remaining constant during state transition.

The \textit{X-axis} position $x$ transition function is described by:
{\small
\begin{equation}
    \begin{split}
    \hat{x}_{t, t-1} &= x_{t-1} + \int_{(t-1)\Delta t}^{t\Delta t} {v_{t-1}}{\cos(\theta(\tau))} d\tau \\
    &= \begin{cases}
    x_{t-1} + \dfrac{v_{t-1}}{\omega_{t-1}}\Bigl[\sin(\theta_{t-1} + \omega_{t-1}\Delta t) \\
    \quad - \sin(\theta_{t-1})\Bigr]. & \omega\neq0 \\
    x_{t-1} + vx_{t-1}\Delta t. & \omega =0
    \end{cases} \\
    \end{split}
    \label{ctrv_x_expand}
\end{equation}
}

The \textit{Y-axis} position $y$ transition function is described by:
{\small
\begin{equation}
    \begin{split}
    \hat{y}_{t, t-1} &= y_{t-1} + \int_{(t-1)\Delta t}^{t\Delta t} {v_{t-1}}{\sin(\theta(\tau))} d\tau \\
    &= \begin{cases}
    y_{t-1} - \dfrac{v_{t-1}}{\omega_{t-1}}\Bigl[\cos(\theta_{t-1} + \omega_{t-1}\Delta t) \\
    \quad - \cos(\theta_{t-1})\Bigr]. & \omega\neq0 \\
    y_{t-1} + vy_{t-1}\Delta t. & \omega =0
    \end{cases} \\
    \end{split}
    \label{ctrv_y_expand}
\end{equation}
}

The heading angle $\theta$ is described as:
\begin{equation}
    \begin{split}
     \hat{\theta}_{t, t-1}&= \theta_{t-1} + \int_{(t-1)\Delta t}^{t\Delta t} {\omega_{t-1}} d\tau  \\
               &= \theta_{t-1} + {\omega_{t-1}} \Delta t.
    \end{split}
    \label{ctrv_y_predict}
\end{equation}

The yaw vector $(\cos\hat{\theta}_{t, t-1}, \sin\hat{\theta}_{t, t-1})$ is the component of $\hat{\theta}_{t, t-1}$.

Velocity $v$ is considered to be constant:
\begin{equation}
    \begin{split}
     \hat{v}_{t, t-1}&= v_{t-1},  \\
    \end{split}
    \label{ctrv_y_predict}
\end{equation}

The \textit{X-axis} velocity $vx$ is described by:
\begin{equation}
    \begin{split}
     \hat{vx}_{t, t-1}&= \hat{v}_{t, t-1}\cos(\theta_{t, t-1}) \\
     &= \hat{v}_{t-1}\cos(\theta_{t-1} + {\omega_{t-1}} \Delta t).
    \end{split}
    \label{ctrv_y_predict}
\end{equation}

The \textit{Y-axis} velocity $vy$ is described by:
\begin{equation}
    \begin{split}
     \hat{vy}_{t, t-1}&= \hat{v}_{t, t-1}\sin(\theta_{t, t-1}), \\
     &= \hat{v}_{t-1}\sin(\theta_{t-1} + {\omega_{t-1}} \Delta t).
    \end{split}
    \label{cv_y_predict}
\end{equation}

\textbf{CTRA.}
\textit{CTRA} is the abbreviation of \textit{Constant Turn Rate and Acceleration}.
The combined acceleration $a_{t-1}$ and yaw rate $w_{t-1}$ are decoded from the instance query $q_{t-1}$, remaining constant during state transition.
The \textit{X-axis} position $x$ transition function is described by:
{\small
\begin{equation}
    \begin{split}
    \hat{x}_{t, t-1} &= x_{t-1} + \int_{(t-1)\Delta t}^{t\Delta t} {v(\tau)}{\cos(\theta(\tau)}) d\tau \\
    &= \begin{cases}
    x_{t-1} + \frac{(v_{t-1} + a_{t-1}\Delta t)\sin(\theta_{t-1}+\omega_{t-1}\Delta t)}{\omega_{t-1}} \\
    - \frac{a_{t-1}\cos(\theta_{t-1})}{\omega_{t-1}^2} + \frac{a_{t-1}\cos(\theta_{t-1}+\omega_{t-1}\Delta t)}{{\omega_{t-1}^2}} \\
    - \frac{v_{t-1}\sin(\theta_{t-1})}{\omega_{t-1}}. & \omega\neq0 \\
    x_{t-1} + (v_{t-1}\Delta t + \frac{a_{t-1}\Delta t^2}{2})\cos(\theta_{t-1}). & \omega =0
    \end{cases}
    \end{split}
    \label{cv_x_predict}
\end{equation}
}

The \textit{Y-axis} position $y$ transition function is described by:
{\small
\begin{equation}
    \begin{split}
    \hat{y}_{t, t-1} &= y_{t-1} + \int_{(t-1)\Delta t}^{t\Delta t} {v(\tau)}{\sin(\theta(\tau))} d\tau \\
        &= \begin{cases}
            y_{t-1} - \frac{(v_{t-1} + a_{t-1}\Delta t)\cos(\theta_{t-1}+\omega_{t-1}\Delta t)}{\omega_{t-1}} \\ 
            - \frac{a_{t-1}\sin(\theta_{t-1})}{\omega_{t-1}^2} + \frac{a_{t-1} \sin(\theta_{t-1}+\omega_{t-1}\Delta t)}{\omega_{t-1}^2} \\
            + \frac{v_{t-1}\cos(\theta_{t-1})}{\omega_{t-1}}. & \omega\neq0 \\
            y_{t-1} + (v_{t-1}\Delta t + \frac{a_{t-1}\Delta t^2}{2})\sin(\theta_{t-1}). & \omega =0
            \end{cases}
    \end{split}
    \label{cv_y_predict}
\end{equation}
}

The combined velocity $v$ is described as:
\begin{equation}
    \begin{split}
     \hat{v}_{t, t-1} &= v_{t-1} + \int_{(t-1)\Delta t}^{t\Delta t} {a} d\tau,  \\
             &= v_{t-1} + {a_{t-1}} \Delta t
    \end{split}
    \label{cv_y_predict}
\end{equation}

The \textit{X-axis} velocity $vx$ is described by:
\begin{equation}
    \begin{split}
     \hat{vx}_{t, t-1}&= \hat{v}_{t, t-1}\cos(\theta_{t, t-1}) \\
     &= (v_{t-1} + {a_{t-1}} \Delta t)\cos(\theta_{t-1} + {\omega_{t-1}} \Delta t).
    \end{split}
    \label{ctrv_y_predict}
\end{equation}

The \textit{Y-axis} velocity $vy$ is described by:
\begin{equation}
    \begin{split}
     \hat{vy}_{t, t-1}&= \hat{v}_{t, t-1}\sin(\theta_{t, t-1}), \\
     &= (v_{t-1} + {a_{t-1}} \Delta t)\sin(\theta_{t-1} + {\omega_{t-1}} \Delta t).
    \end{split}
    \label{cv_y_predict}
\end{equation}

The heading angle $\theta$ is described as:
\begin{equation}
    \begin{split}
     \hat{\theta}_{t, t-1}&= \theta_{t-1} + \int_{(t-1)\Delta t}^{t\Delta t} {\omega_{t-1}} d\tau  \\
             &= \theta_{t-1} + {\omega_{t-1}} \Delta t.
    \end{split}
    \label{cv_y_predict}
\end{equation}

The yaw vector $(\cos\hat{\theta}_{t, t-1}, \sin\hat{\theta}_{t, t-1})$ is the component of $\hat{\theta}_{t, t-1}$.

\label{sec:app_screen}
\begin{figure*}[t!]
  \centering
  \includegraphics[width=0.9\linewidth]{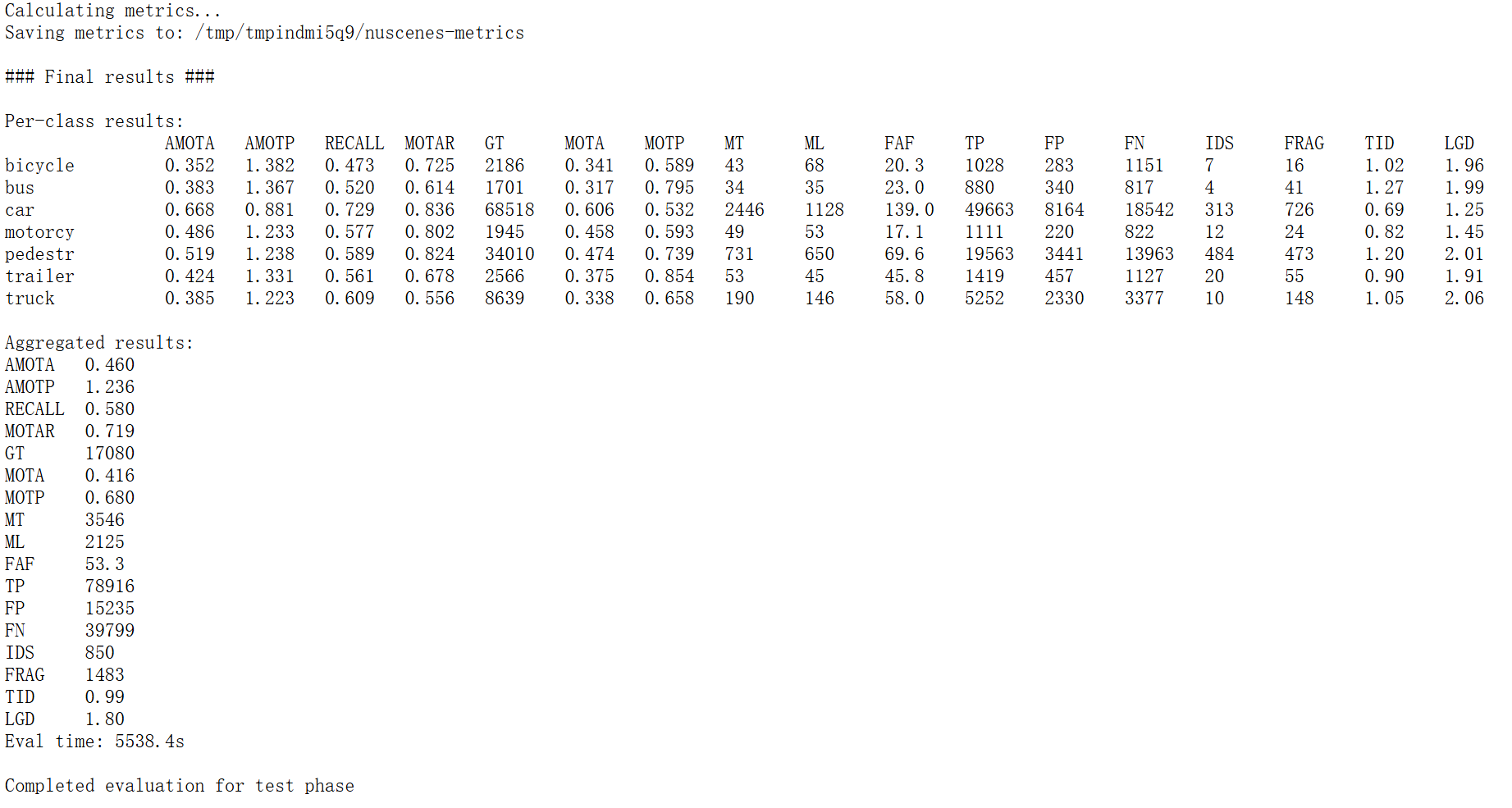}
  \caption[l]{The screenshot of our tracking result (ADA-Track~\cite{ding2024ada}-HAT) on the nuScenes test set.
  }
\label{fig:ada_track_hat_test}
\end{figure*}

\paragraph{The Stability of HAT.}
\label{sec:stability_hat}
In this section, we prove the stability of our proposed HAT module from the perspective of anchor decoding.
We define $\tilde{X}_{t, t-1} \in \mathbb{R}^{M \times 3}$ as the position hypotheses of a instance $\tilde{b}_{t, t-1}$, with corresponding weights $w_{a} \in \mathbb{R}^{M \times 1}$ in $W_{a}$.
Notably, $w_{a}$ is positive, and the sum of $w_{a}$ is constrained to 1 by the softmax function.
The weighted summation of $\tilde{X}_{t, t-1}$ is expressed as:
{\small
\begin{equation}
\begin{aligned}
(\bar{X}_{t, t-1})^{T} &= (\sum_{m=1}^{M}w^{m}_{a}\tilde{X}^{m}_{t, t-1})^{T} \\
  &= (R^{t}_{t-1}X^{T}_{t, t-1} + T^{t}_{t-1}) + R^{t}_{t-1}\sum_{m=1}^{M}w^{m}_{a}(\Delta X^{m})^{T}, \\
  \bar{X}_{t, t-1} &= \tilde{X}_{t-1} + \Delta \tilde{X}.
\label{linear_sum}
\end{aligned}
\end{equation}
}

$\tilde{X}_{t-1}$ represents the warped position of historical anchor, and $\bar{X}_{t, t-1}$ denotes the decoded position.
The term $\Delta \tilde{X}$ represents the adaptive motion compensation, constrained within
$(R^{t}_{t-1}\min(\Delta X^{m}), R^{t}_{t-1}\max(\Delta X^{m}))$.
$\Delta X^{m}$ is the position compensation of the $i$-th motion model.
Since these models are well-established, this constraint ensures that $\bar{X}_{t, t-1}$ remains stable, guaranteeing a lower bound on convergence and accuracy.

\subsection{Additional Experiments}



\begin{table}[h]
  \centering

  {
  \tabcolsep=1pt
  \begin{tabular}{c|cc|ccc|c}
    \toprule
    & \multicolumn{2}{c|}{Temporal} & \multicolumn{3}{c|}{Spatial} & \multirow{2}{*}{Overall} \\
    \cmidrule(lr){2-3} \cmidrule(lr){4-6}
    & Anchor & Feat & Anchor & Feat & Mixing & \\
    \midrule
    Latency & 3.1 & 1.1 & 0.6 & 0.4 & 2.2 & 7.4 \\
    \bottomrule
  \end{tabular}
  }
  \caption{Latency (ms) of HAT module in SparseDrive-HAT on the nuScenes validation set with 600 cached instances.}
  \label{tab:runtim}
\end{table}

\begin{table}[t]
  \centering
  \small
  \begin{center}
  {
    \tabcolsep=1.4pt
    \begin{tabular}{c|c|c|ccccc}
      \toprule
      Detector        & NDS$\uparrow$    & mAP$\uparrow$         & mATE$\downarrow$    & mAOE$\downarrow$ & mAVE$\downarrow$ \\
      \midrule
      Baseline         & 74.8    & 73.0                         & 0.2808             & 0.2502             & 0.2049            \\
    Baseline-HAT            &  \textbf{75.4 \textit{{\scriptsize (+0.6)}}}   &  \textbf{73.6 \textit{{\scriptsize (+0.6)}}}                     & \textbf{0.2794}             & \textbf{0.2395}             & \textbf{0.1805}          \\
      \bottomrule
       
    \end{tabular}
  }
  \end{center}
  \caption{Comparison results on nuScenes validation set for LiDAR-Camera 3D temporal detector MV2DFusion.}
  \label{tab:append_mv2d}
\end{table}

\paragraph{Computational Cost of HAT.}
As illustrated in~\cref{tab:runtim}, we evaluate the latency of each HAT component on SparseDrive~\cite{sun2024sparsedrive} using an RTX 3090.
The method caches 600 historical instances for temporal enhancement. 
Compared to the baseline latency of 111ms, HAT adds only 7.4ms (increased by 6.7\%) overhead and reduces collisions by 30\%, confirming practical deployability.
The anchor generator is the main bottleneck, while fixed motion models allow acceleration via dedicated operators.

\paragraph{Performance of HAT on LiDAR-Camera Detector.}
As depicted in~\cref{tab:append_mv2d}, we further validate the effectiveness of HAT on the state-of-the-art LiDAR-Camera 3D temporal detector MV2DFusion~\cite{wang2024mv2dfusion}. 
HAT exhibits strong generalization capability, consistently improving performance by 0.6\% in both NDS and mAP over the baseline.

      

\subsection{Performance Verification}

As shown in \cref{fig:ada_track_hat_test}, we present the tracking performance of ADA-Track~\cite{ding2024ada} integrated with our proposed HAT module on the nuScenes test set.

\begin{figure*}[t!]
  \centering
  \includegraphics[width=1\linewidth]{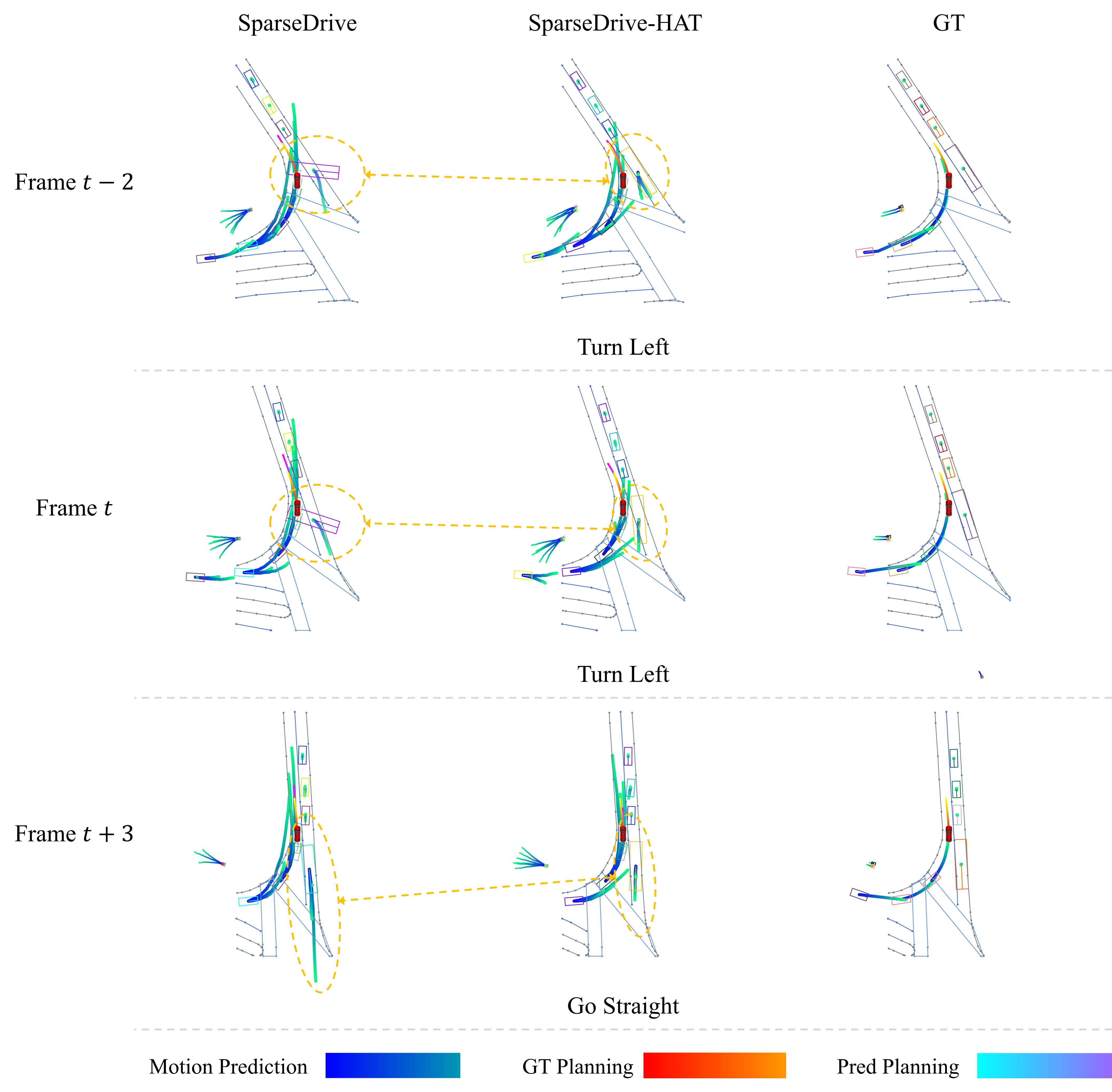}
  \caption[l]{Visualization results of E2E AD methods SparseDrive and SparseDrive-HAT in the turning scenario from the nuScenes dataset. The official visualization script provided by the baseline is used.
  }
\label{fig:e2e_viz}
\end{figure*}

\subsection{Additional Qualitative Results}

As shown in~\cref{fig:e2e_viz}, embedding HAT enhances perception of baseline~\cite{sun2024sparsedrive} by reducing the IDS and velocity/heading errors for the key obstacle (highlighted bus), thereby increasing perceptual reliability.
This enhanced perception enables SparseDrive-HAT to exhibit more cautious behavior (frame $t-2$ and $t$), generating a safer planning trajectory with arc-shaped obstacle avoidance that better aligns with the expert trajectory.
Together with the quantitative results in~\cref{tab:test_ad}, these qualitative findings demonstrate the effectiveness of HAT in end-to-end autonomous driving systems and underscore the importance of accurate motion modeling.

\clearpage


{
    \small
    \bibliography{aaai2026}
}

\end{document}